\pdfoutput=1

\documentclass[11pt]{article}
\usepackage[preprint]{acl}

\usepackage{times}
\usepackage{latexsym}

\usepackage[T1]{fontenc}

\usepackage[utf8]{inputenc}

\usepackage{microtype}

\usepackage{inconsolata}

\usepackage{multirow}
\usepackage{graphicx}
\usepackage{booktabs}
\usepackage{algorithm}
\usepackage{algpseudocode}
\usepackage{amsmath}
\usepackage{amsfonts}
\usepackage{bm}
\usepackage[normalem]{ulem}
\usepackage{tabularx}
\usepackage{mathabx}  
\usepackage{colortbl}  
\usepackage{hyperref}
\usepackage{enumitem}
\usepackage{tikz}

\useunder{\uline}{\ul}{}

\def\eqref#1{equation~\ref{#1}}

\def\1{\bm{1}}

\def\va{{\bm{a}}}

\def\vc{{\bm{c}}}

\def\vp{{\bm{p}}}

\def\evp{{p}}

\DeclareMathAlphabet{\mathsfit}{\encodingdefault}{\sfdefault}{m}{sl}
\SetMathAlphabet{\mathsfit}{bold}{\encodingdefault}{\sfdefault}{bx}{n}

\newcommand{\R}{\mathbb{R}}

\definecolor{Gray}{gray}{0.95}

\newcommand{\ours}[0]{Beam Aggregation}

\newcommand{\oursbf}[0]{\textbf{\ours{}}}

\newcommand{\LLM}[0]{\mathrm{LLM}}

\title{\textbf{BeamAggR}: Beam Aggregation Reasoning over Multi-source Knowledge for Multi-hop Question Answering}

\author{
    Zheng Chu$^{1}$, 
    Jingchang Chen$^{1}$, 
    Qianglong Chen$^{2}$\footnotemark[1],
    Haotian Wang$^{1}$ \\
    \textbf{
    Kun Zhu$^{1}$, 
    Xiyuan Du$^{1}$, 
    Weijiang Yu$^{3}$},
    \textbf{
    Ming Liu$^{1,4}$\thanks{$\,$Corresponding Authors: Ming Liu, Qianglong Chen}, 
    Bing Qin$^{1,4}$
    } \\
    $^{1}$Harbin Institute of Technology, Harbin, China \\
    $^{2}$Zhejiang University~~
    $^{3}$Sun Yat-sen University~~
    $^{4}$Peng Cheng Laboratory~~ \\
    \texttt{\{zchu,jcchen,mliu\}@ir.hit.edu.cn},
    \texttt{chenqianglong.ai@gmail.com}
}

\definecolor{bluegray}{HTML}{647D87}
\definecolor{Azure1}{HTML}{F0FFFF}
\definecolor{Burlywood4}{HTML}{8B7355}

\newcommand{\promptfigure}[2]{
\begin{tikzpicture}[
    every node/.style={outer sep=0},
    window/.style={rectangle, draw=black, rounded corners, thick, align=left, font=\small, inner xsep=0pt, inner ysep=6mm},
    titlestyle/.style={text=white},
]
\node[window] (window) at (0,-0.2cm) [minimum width=\linewidth, text width=\linewidth-6mm, anchor=north] {#2};
\draw[draw=black, fill=bluegray, thick] (-\linewidth/2,-0.35cm)
    [sharp corners] -- (\linewidth/2, -0.35cm)
    [rounded corners=5pt] -- (\linewidth/2, 0.35cm)
    [rounded corners=5pt] -- (-\linewidth/2, 0.35cm)
    [sharp corners] -- cycle;
\node[titlestyle] (title) at (0,-0.4mm) {#1};
\end{tikzpicture}%
}

\newcommand{\examplefigure}[2]{
\begin{tikzpicture}[
    every node/.style={outer sep=0},
    window/.style={rectangle, draw=black, rounded corners, thick, align=left, font=\small, inner xsep=0pt, inner ysep=6mm},
    titlestyle/.style={text=white},
]
\node[window] (window) at (0,-0.2cm) [minimum width=\linewidth, text width=\linewidth-6mm, anchor=north] {#2};
\draw[draw=black, fill=Burlywood4, thick] (-\linewidth/2,-0.35cm)
    [sharp corners] -- (\linewidth/2, -0.35cm)
    [rounded corners=5pt] -- (\linewidth/2, 0.35cm)
    [rounded corners=5pt] -- (-\linewidth/2, 0.35cm)
    [sharp corners] -- cycle;
\node[titlestyle] (title) at (0,-0.4mm) {#1};
\end{tikzpicture}%
}
\makeatletter
\def\therule{\makebox[\algorithmicindent][l]{\hspace*{.5em}\vrule height .75\baselineskip depth .25\baselineskip}}%

\newtoks\therules
\therules={}
\def\appendto#1#2{\expandafter#1\expandafter{\the#1#2}}
\def\gobblefirst#1{
  #1\expandafter\expandafter\expandafter{\expandafter\@gobble\the#1}}%
\def\LState{\State\unskip\the\therules}
\def\pushindent{\appendto\therules\therule}%
\def\popindent{\gobblefirst\therules}%
\def\printindent{\unskip\the\therules}%
\def\printandpush{\printindent\pushindent}%
\def\popandprint{\popindent\printindent}%

\algdef{SE}[WHILE]{While}{EndWhile}[1]
  {\printandpush\algorithmicwhile\ #1\ \algorithmicdo}
  {\popandprint\algorithmicend\ \algorithmicwhile}%
\algdef{SE}[FOR]{For}{EndFor}[1]
  {\printandpush\algorithmicfor\ #1\ \algorithmicdo}
  {\popandprint\algorithmicend\ \algorithmicfor}%
\algdef{S}[FOR]{ForAll}[1]
  {\printindent\algorithmicforall\ #1\ \algorithmicdo}%
\algdef{SE}[LOOP]{Loop}{EndLoop}
  {\printandpush\algorithmicloop}
  {\popandprint\algorithmicend\ \algorithmicloop}%
\algdef{SE}[REPEAT]{Repeat}{Until}
  {\printandpush\algorithmicrepeat}[1]
  {\popandprint\algorithmicuntil\ #1}%
\algdef{SE}[IF]{If}{EndIf}[1]
  {\printandpush\algorithmicif\ #1\ \algorithmicthen}
  {\popandprint\algorithmicend\ \algorithmicif}%
\algdef{C}[IF]{IF}{ElsIf}[1]
  {\popandprint\pushindent\algorithmicelse\ \algorithmicif\ #1\ \algorithmicthen}%
\algdef{Ce}[ELSE]{IF}{Else}{EndIf}
  {\popandprint\pushindent\algorithmicelse}%
\algdef{SE}[PROCEDURE]{Procedure}{EndProcedure}[2]
   {\printandpush\algorithmicprocedure\ \textproc{#1}\ifthenelse{\equal{#2}{}}{}{(#2)}}%
   {\popandprint\algorithmicend\ \algorithmicprocedure}%
\algdef{SE}[FUNCTION]{Function}{EndFunction}[2]
   {\printandpush\algorithmicfunction\ \textproc{#1}\ifthenelse{\equal{#2}{}}{}{(#2)}}%
   {\popandprint\algorithmicend\ \algorithmicfunction}%
\makeatother

\begin{document}
\maketitle

\begin{abstract}
Large language models (LLMs) have demonstrated strong reasoning capabilities.
Nevertheless, they still suffer from factual errors when tackling knowledge-intensive tasks.
Retrieval-augmented reasoning represents a promising approach.
However, significant challenges still persist, including inaccurate and insufficient retrieval for complex questions, as well as difficulty in integrating multi-source knowledge.
To address this, we propose Beam Aggregation Reasoning (\textbf{BeamAggR}), a reasoning framework for knowledge-intensive multi-hop QA.
BeamAggR explores and prioritizes promising answers at each hop of question.
Concretely, we parse the complex questions into trees, which include atom and composite questions, followed by bottom-up reasoning.
For atomic questions, the LLM conducts reasoning on multi-source knowledge to get answer candidates.
For composite questions, the LLM combines beam candidates, explores multiple reasoning paths through probabilistic aggregation, and prioritizes the most promising trajectory.
Extensive experiments on four open-domain multi-hop reasoning datasets show that our method significantly outperforms SOTA methods by 8.5\%.
Furthermore, our analysis reveals that BeamAggR elicits better knowledge collaboration and answer aggregation.

\end{abstract}
\section{Introduction}
\label{sec:introduction}

Large language models have showcased impressive performance across various NLP tasks~\citep{gpt-4,llama2}.
Furthermore, chain-of-thought (CoT) prompting further enhances the reasoning capabilities of LLMs~\citep{fewshot-cot,zeroshotcot,survey-cot}.
However, when the question surpasses the knowledge boundaries of LLMs, it leads to factual errors, also known as hallucination.
Retrieval-augmented generation (RAG) assists LLMs by retrieving supporting knowledge to alleviate factual errors,
thereby drawing significant attention within the research community~\citep{rag-survey,hallucination-survey}.

Early attempts adopt a \textit{retrieve-then-read} framework for one-time retrieval~\citep{oner-internellm,oner-retro,oner-atlas,mhqa-iag}.
They use the original complex question as the retrieval query and achieve satisfactory performance in single-hop questions~\citep{tqa,nq}.
Nonetheless, when confronted with complex multi-hop questions, one-time retrieval suffers from the issues of inaccurate and irrelevant retrieval, which greatly impair their performance in multi-hop reasoning.

\begin{figure}[t]
    \centering
    \includegraphics[clip, width=\linewidth]{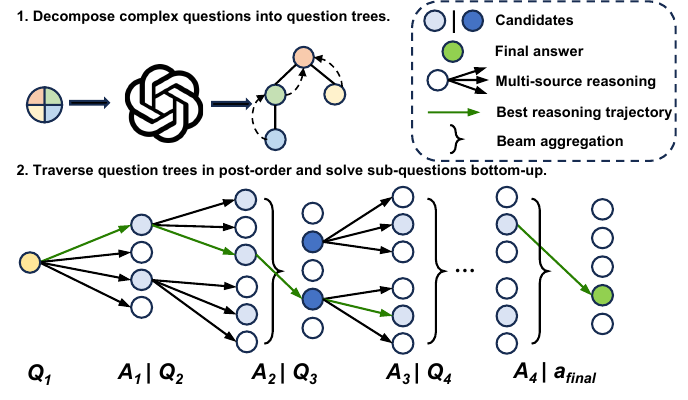}
    \caption{
A brief overview of our method.
Complex questions are decomposed into trees (top).
Multi-source beam aggregation reasoning is conducted to find the best reasoning trajectory. (bottom)
}
    \label{fig:intro}
\end{figure}

Recent work introduces iterative multi-round retrieval~\citep{mhqa-dsp,bamboogle-selfask,mhqa-ircot,mhqa-flare,mhqa-iterretgen}.
They use the content generated by LLMs for retrieval and, in turn, use the newly retrieved content for reasoning.
Through the iterative alternation between retrieval-augmented reasoning and reasoning-augmented retrieval, the retrieval is substantially improved.
Meanwhile, a portion of research decomposes complex questions into simple sub-questions, employing sub-question retrieval to obtain more precise information~\citep{least2most,mhqa-ptot,mhqa-graphguided,mhqa-semistruct}.

However, there are still significant issues with these methods.
Iterative retrieval is tough to achieve precise retrieval aligned with the model's reasoning.
Sub-question retrieval grapples with the challenge of accurately aggregating answers, which causes cascading errors.
Besides, when it comes to the open-domain setting, relying solely on knowledge from a single source proves inadequate for complex questions.
Introducing multi-source knowledge may encounter knowledge conflicts, rendering efficient collaboration challenging, thus impeding its applications.

To address the aforementioned challenges, our research focuses on the following question: 
\textit{How can models adaptively select and integrate knowledge from different sources during the reasoning, while reducing cascading errors in sub-questions aggregation?}
Building upon this motivation, we propose \ours{} Reasoning (\textbf{BeamAggR}) framework for knowledge-intensive multi-hop reasoning.
Concretely, our method consists of three modules.
(i) \textit{question decomposition}:
We begin by leveraging LLMs to decompose complex questions and convert them into question trees.
The root node contains the original complex question, leaf nodes contain atomic sub-questions, and intermediate nodes contain composite questions that require compositional (or comparative) reasoning to obtain answers.
Afterward, we traverse the question tree in post-order, employing bottom-up reasoning. 
(ii) \textit{complementary multi-source reasoning}:
For atomic questions, we conduct multi-source reasoning, followed by fine-grained answer aggregation, thus fostering knowledge collaboration.
The aggregated answers are then normalized into a probability distribution, serving as candidates for beam aggregation.
(iii) \textit{beam aggregation}:
For composite questions, we enumerate the combinations of their dependent sub-questions and conduct reasoning.
Finally, the reasoning results are probabilistically aggregated, and the most promising predictions are selected.

In summary, 
our method explores reasoning trajectories at each hop of questions and prioritizes paths with higher likelihoods, thereby bringing reasoning insight and reducing cascading errors.
Besides, the complementary multi-source reasoning helps alleviate knowledge omission and conflict.

We evaluate our method on four open-domain multi-hop reasoning datasets: HotpotQA~\citep{hotpotqa}, 2WikiMQA~\citep{2wikimhqa}, MuSiQue~\citep{musique}, and Bamboogle~\citep{bamboogle-selfask}.
The experiments are conducted using GPT-3.5-turbo~\citep{gpt35} and Mistral~\citep{mistral}.
Experimental results show that our method significantly outperforms the baselines on four datasets, with an average improvement of 8.5\% compared to the previous state-of-the-art method, demonstrating its superiority.
Furthermore, thorough analysis reveals the superiority of our approach to knowledge collaboration and answer aggregation.

\noindent Our contributions can be summarized as follows:
\begin{itemize}[itemsep=2pt,topsep=2pt,parsep=0pt,leftmargin=*]
    \item We introduce BeamAggR, a framework for open-domain multi-hop reasoning, which outperforms the state-of-the-art methods.
    \item BeamAggR dynamically integrates multi-source knowledge in fine granularity during reasoning, fostering knowledge collaboration.
    \item BeamAggR broadens the scope of reasoning with beam combination and optimizes reasoning trajectories, mitigating cascading errors.
\end{itemize}

\section{Related Work}
\label{sec:related_work}

\begin{figure*}[t]
    \centering
    \includegraphics[clip, width=\linewidth]{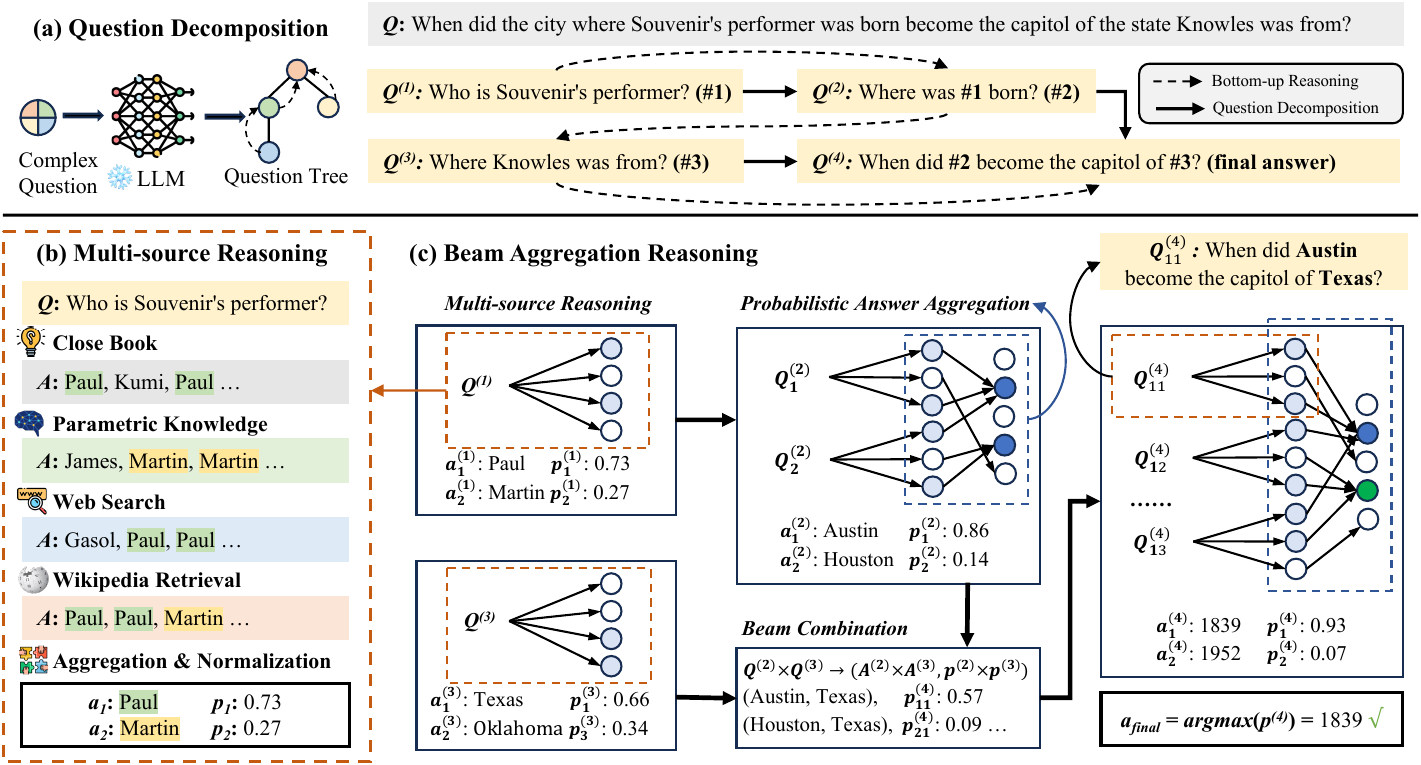}
    \caption{
    An overview of BeamAggR.
    (a) Question decomposition: decompose complex questions into trees and address them bottom-up
    (b) Multi-source reasoning: reason from diverse knowledge sources and normalize answers into a probability distribution
    (c) Beam aggregation reasoning: explore based on children's predictions, probabilistic aggregate answers and select the most promising reasoning trajectory.
    }
\label{fig:method}
\end{figure*}

\subsection{Reasoning with Large Language Model}
\citet{fewshot-cot} prompts LLMs to generate reasoning process before final answers, which is known as chain-of-thought (CoT) prompting.
Since then, CoT prompting has been widely applied to enhance the reasoning capabilities of LLMs.
Some work also designs instructions or clustering demonstrations for zero-shot reasoning~\citep{zeroshotcot,autocot}.
Additionally, self-ensemble has been proven through extensive experiments to be an effective approach to improve performance. 
\citet{self-consisntency} uses probabilistic sampling for multiple reasoning traces, 
while \citet{crosslingualcot} diversifies reasoning paths by using multiple languages CoTs.
To address complex questions, \citet{least2most,successiveprompting} decompose them into sub-questions and solve them progressively,
while \citet{tot} models reasoning procedures as BFS or DFS search on reasoning trees.

In contrast, our method employs divide-and-conquer strategy, breaking down complex questions into trees and addressing them through bottom-up aggregation reasoning.

\subsection{Retrieval-augmented Reasoning}
Knowledge-sensitive tasks may induce factual hallucinations in LLMs, thus necessitating external retrieval for retrieval-augmented reasoning.
Early work uses one-time retrieval, but they struggle to gather all the necessary knowledge to answer complex questions, resulting in knowledge omissions~\citep{oner-retro,oner-internellm,oner-atlas}.
To address this, iterative retrieval is proposed.
DSP~\citep{mhqa-dsp} engages in iterative interactions between retriever and reader through programmatically defined procedures.
SelfAsk~\citep{bamboogle-selfask} iteratively decomposes questions and solves them through the Google search.
IRCoT~\citep{mhqa-ircot} uses each reasoning step as a query for retrieval until obtaining the final answer.
Similarly, ITER-RETGEN~\citep{mhqa-iterretgen} conducts iterative retrievals by concatenating the output from the previous round with the original question. 
FLARE~\citep{mhqa-flare} introduces a lookahead mechanism, dynamically controlling the timing of retrieval based on reasoning confidence.
Beam Retrieval~\citep{beam-retrieval} introduces an end-to-end framework designed to retrieve relevant paragraphs at each hop of questions through beam search.
Meanwhile, some efforts achieve more precise retrieval by decomposing the problem into QDMR format~\citep{missref-mhqa1}.
\citet{mhqa-ptot} decomposes the problem into a tree, while \citet{mhqa-graphguided} constructs a reasoning graph.

Compared to their approach, our method integrates diverse knowledge via complementary multi-source reasoning.
Besides, it explores reasoning trajectories at each hop of questions and prioritizes the most promising path, thereby eliciting better answer aggregation and reducing cascading errors.

\section{Beam Aggregation Reasoning}
\label{sec:method}

Beam Aggregation Reasoning decomposes complex questions into trees and conducts bottom-up reasoning.
Throughout the bottom-up reasoning, it intricately amalgamates diverse knowledge sources, thereby mitigating the dearth of knowledge.
Furthermore, beam aggregation probabilistically consolidates answers and selects reasoning pathways, consequently diminishing cascading errors.

Firstly, we decompose complex questions, then perform bottom-up multi-source knowledge reasoning in a post-order traversal manner, and employ beam aggregation at intermediate nodes until reaching the root node to obtain the final answer.
The procedure is depicted in Algorithm~\ref{alg:ours}, with an overview of the methodology illustrated in Figure~\ref{fig:method}.
We will introduce question decomposition in \S~\ref{subsec:question_decomposition}, 
multi-source reasoning in \S~\ref{subsec:reader} and beam aggregation in \S~\ref{subsec:beam_aggregation}.
Detailed definitions of notations in the algorithm and formulas can be found in Table~\ref{tab:symbol}.
Task definition is given in Appendix~\ref{appendix:task_def}.

\begin{algorithm}[t]
\begin{algorithmic}[1]
\Require Complex multi-hop questions, $Q$
\Require Multi-source knowledge retriever, $R$
\Require Large language model, $LLM$
\Require Candidate size in aggregation, $k$
\State $Q_{decomp} = LLM(Q)$
\For{$N^{(i)}$ in $\textrm{PostOrderTraverse}(Q_{decomp})$}
    \If{$N^{(i)}$ is leaf-node}

        \LState $\hat{\va} = LLM(q^{(i)}, ~R)$
        \LState $\va^{(i)}, ~\vp^{(i)} = \textrm{Vote}(\hat{\va})[1:k]$
    \Else
        \LState \textbf{initialize} $\hat{\va}, ~\hat{\vp} = list, ~list$ 
        \For{$c', ~p'$ in $\textrm{CartProd}(\textrm{sons}(N^{(i)}))$}
            \LState $q' = \textrm{MaskFill}(q_i, ~c')$
            \LState $\va^{t}, \vp^{t} = \textrm{Vote}(LLM(q', ~c', ~R))$
            \LState $\hat{\va} \gets \textrm{Concat}(\hat{\va}, ~\va^{t})$
            \LState $\hat{\vp} \gets \textrm{Concat}(\hat{\vp}, ~\vp^{t} \cdot p')$
        \EndFor
        \LState $\va^{(i)}, ~\vp^{(i)} = \textrm{Aggr}(\hat{\va}, ~\hat{\vp})[1:k]$
    \EndIf
\EndFor
\\ \Return $\va^{root}\textrm{[1]}$
\end{algorithmic} 
\caption{\ours{} Reasoning}
\label{alg:ours} 
\end{algorithm}

\subsection{Question Decomposition}
\label{subsec:question_decomposition}

Multi-hop questions entail complex structures, such as bridge, comparison, composition and their integration.
To address this, we parse complex questions into trees to express the reasoning structure.
As shown in Figure~\ref{fig:method} (a), the complex question \textit{Q} is decomposed into a tree comprising four simpler sub-questions, $Q^{(1)}$ to $Q^{(4)}$, with (compositional) dependencies among them.
The root node represents the original complex question, the leaf nodes represent atomic sub-questions, and the intermediate nodes require compositional (comparative) reasoning to obtain answers.
Following~\citet{mhqa-ptot,mhqa-semistruct}, we adopt \#i as the placeholder for intermediate questions, enabling us to replace incomplete questions with solved sub-questions as we traverse to that node.
Specifically, the decomposed question is represented in QDMR format~\cite{missref-mhqa1}.
Afterward, we tackle the complex questions in a bottom-up fashion, following a post-order traversal sequence.
The tree is represented as a post-order traversal node sequence, $Q_{decomp}=\{N^{(1)}, N^{(2)} \dots\}$, with each node containing a question, candidate answers, and the associated probabilities, $N^{(i)}$ = $\{q^{(i)}, ~\va^{(i)}, ~\vp^{(i)}\}$.
It is noteworthy that the model may produce structural incorrect decomposition, which needs post-filtering to ensure the validity of decomposition.
We present the formal definitions in Table~\ref{tab:symbol}(b).

\subsection{Complementary Multi-source Reasoning}
\label{subsec:reader}
To avoid the hallucination caused by lack of knowledge, we use four reasoning strategies combined with answer normalization to fuse information from diverse knowledge sources, as shown in Figure~\ref{fig:method}(b).
The knowledge sources include implicit internal knowledge, explicit internal knowledge, Wikipedia knowledge, and web search knowledge.
\begin{align}
    \va_{s} = \LLM{}(q, K_{s}) 
\end{align}
where $s \in \{closebook, parametric, wiki, serp\}$ is reasoning strategy, $K$ is retrieved knowledge.

\paragraph{Internal Knowledge Reasoning}
For implicit knowledge reasoning, 
we prompt the model with chain-of-thought demonstrations to perform closed-book reasoning.
There are also studies suggesting that the model's parametric knowledge can serve as a retrieval source~\citep{internal-genread,mhqa-iag}.
We first prompt the model to generate parametric knowledge relevant to the question, and then employ it for explicit knowledge reasoning.

\paragraph{External Knowledge Reasoning}
We utilize Wikipedia and search engines as external retrieval for external knowledge reasoning.
Regarding Wikipedia, we employ BM25 to conduct sparse retrieval over the full Wikipedia dumps. 
For search engines, we call the Google Search API and use the organic results as the retrieval content.
After retrieval, we employ few-shot prompts to conduct reasoning on the retrieved documents.

\paragraph{Answer Normalization}
After completing four sets of independent knowledge reasoning, we merge them through voting to achieve knowledge fusion.
Following that, we normalize the answer-frequency pairs to probability distributions for subsequent beam aggregation, as shown in Eq.~(\ref{equ:normalize}).
\begin{align}
    \evp_{i} = \frac
    {\mathrm{exp}(f_{i} / \tau)}
    {\sum_{j=1}^{k} {\mathrm{exp}(f_{j} / \tau)}}
\label{equ:normalize}
\end{align}
where $f$ is frequency and $\tau$ is temperature.

\subsection{Beam Aggregation}
\label{subsec:beam_aggregation}
In beam aggregation, we conduct reasoning over combinations of candidate answers inherited from sub-questions to expand reasoning breadth and exploration space, as shown in Figure~\ref{fig:method}(c).
We then select reasoning paths by maximizing the marginal probability distribution of predictions.

\paragraph{Beam Combination}
In intermediate nodes, we need to conduct reasoning based on the answers derived from previous sub-questions,
and each sub-question is associated with a set of answers and probabilities, termed candidates (or beams).
The intermediate question may depend on multiple sub-questions, so we enumerate combinations among the candidates.
Specifically, we compute the Cartesian product among the candidates,
as shown in Eq.~(\ref{equ:cartprod}).
To prevent combinatorial explosion, we restrict the exploration space with beam size.
Afterward, for each combination, we substitute the placeholders in the question with candidate answers and conduct multi-source reasoning. 
We take a composition node with two sub-questions as an example.
\begin{align}
    \{ \langle a^{(x)}_i, a^{(y)}_j \rangle, p^{(x)}_i p^{(y)}_j ~|~ i, j = 1, 2, \dots, k \}
\label{equ:cartprod}
\end{align}
where $^{(x),(y)}$ denote sub-questions, $k$ is candidate size, $a$ is answer and $p$ is associated probability.

As illustrated in Figure~\ref{fig:method}(c), $Q^{(4)}$ relies on $Q^{(2)}, Q^{(3)}$.
We calculate the Cartesian product of candidates of $Q^{(2)}$ and $Q^{(3)}$, and perform substitution to derive new questions.
For example, $Q^{(4)}_{11}$ is obtained by substitute the \#1 with $a^{(2)}_{1}$ and \#2 with $a^{(3)}_{1}$, “\textit{When did \textbf{Austin} become the capital of \textbf{Texas}?}”,
with $P(Q^{(4)}_{11})=p^{(2)}_{1} p^{(3)}_{1}$.
Subsequently, we conduct multi-source reasoning on all substituted sub-questions.

\paragraph{Probabilistic Answer Aggregation}
We have explored numerous combinations of sub-questions in beam combination, yielding multiple sets of answers and probabilities.
Next, we will aggregate the above answers according to the probabilities to determine the optimal reasoning path.
This can be formalized as argmax for maximizing the marginal probabilities of predictions, as described below.
\begin{align}
    P(y) =& \sum_{q_i \in Q}{P(y | x = q_i)} \cdot P(q_i)
\end{align}
where $Q$ is the original intermediate question, $q_i$ is a substituted question, $P(y|x=q_i)$ is the answers distribution of $q_i$, and $P(q_i)$ is the weight of $q_i$.

After probabilistic answer aggregation, we keep the top-$k$ answers $\va$ and their probabilities $\vp$ as candidates.
The aggregated candidates will continue propagating bottom-up, participating in subsequent beam aggregations until reaching the root node.
Upon reaching the root node, we regard the answer with the highest probability as the final answer.

\section{Experimental Setup}
\label{sec:setup}

\subsection{Datasets}
\label{subsec:datasets}
We evaluate BeamAggR on four open-domain multi-hop reasoning datasets. 
HotpotQA~\citep{hotpotqa}, 2WikiMQA~\citep{2wikimhqa}, and Bamboogle~\citep{bamboogle-selfask} consist of two-hop questions, and MuSiQue~\citep{musique} contains questions with 2 to 4 hops.
For HotpotQA, 2WikiMQA, and MuSiQue, we use the same development and test set provided by IRCoT~\citep{mhqa-ircot}. 
The test set and development set are both extracted from the original development set, consisting of 500 and 100 instances, respectively.
For Bamboogle, we use all 125 instances as the test set, without a separate development set, and use the same hyperparameters with 2WikiMQA.
The evaluation metric is token-level F1.

On all four datasets, we conduct experiments in the open-domain setting, employing the entire Wikipedia dumps and web search for retrieval.

\newcommand{\floatingtext}[1]{\kern-0.2em\makebox[0pt][l]{#1}}
\newcommand{\up}[1]{\floatingtext{\scriptsize($\uparrow$#1)}}

\begin{table*}[t]
\centering
\setlength\tabcolsep{4.5pt}
\resizebox{\textwidth}{!}{%
\begin{tabular}{lccccccccccccc}
\toprule
\multirow{2.5}{*}{\textbf{Methods}} & \multicolumn{3}{c}{\textbf{HotpotQA}} & \multicolumn{4}{c}{\textbf{MuSiQue}} & \multicolumn{5}{c}{\textbf{2WikiMQA}} & \textbf{Bamboogle} \\
\cmidrule(lr){2-4}  \cmidrule(lr){5-8} \cmidrule(lr){9-13} \cmidrule(lr){14-14} 
 & \textbf{Overall} & \textit{Bridge} & \textit{Comp.} & \textbf{Overall} & \textit{2hop} & \textit{3hop} & \textit{4hop} & \textbf{Overall} & \textit{Bridge} & \textit{Infer.} & \textit{Comp.} & \textit{B.C.} & \textbf{Overall} \\
\midrule
\multicolumn{14}{l}{\textbf{\textit{Close-book Reasoning}}} \\
SP & 38.9 & 37.5 & 45.3 & 15.6 & 16.4 & 16.2 & 12.6 & 33.9 & 13.9 & 23.9 & 53.3 & 57.0 & 27.8 \\
CoT & 46.5 & 44.6 & 55.5 & 24.7 & 30.2 & 22.5 & 13.2 & 42.3 & 25.7 & 25.1 & 58.0 & 68.5 & 53.6 \\
\midrule
\multicolumn{14}{l}{\textbf{\textit{Retrieval-augmented Reasoning}}} \\
OneR~$^{\diamondsuit}$ & 55.3 & 52.9 & 66.5 & 16.4 & 22.1 & 10.6 & 10.4 & 42.9 & 24.3 & 28.7 & 75.7 & 51.2 & 46.8 \\
Self-Ask~$^\spadesuit$ & 49.4 & 45.3 & 68.6 & 16.2 & 24.4 & 8.8 & 7.5 & 46.9 & 31.6 & 40.5 & 71.5 & 52.6 & 51.9 \\
IRCoT~$^\spadesuit$ & 56.2 & 53.4 & {\ul 69.6} & 24.9 & 31.4 & 19.2 & {\ul 16.4} & 56.8 & 44.2 & 22.7 & 89.0 & 69.4 & 55.0 \\
FLARE~$^\spadesuit$ & 56.1 & 54.2 & 64.4 & 31.9 & 40.9 & 27.1 & 15.0 & 60.1 & 46.2 & 54.5 & 81.4 & 66.3 & 58.1 \\
ProbTree~$^\clubsuit$ & {\ul 60.4} & {\ul 59.2} & 65.9 & {\ul 32.9} & {\ul 41.2} & {\ul 30.9} & 14.4 & {\ul 67.9} & {\ul 49.8} & \textbf{66.4} & {\ul 81.7} & \textbf{91.1} & {\ul 66.6} \\
\midrule

\textbf{Ours} & \textbf{62.9} \up{2.5} & \textbf{60.5} & \textbf{74.2} & \textbf{36.9} \up{4.0}& \textbf{43.3} & \textbf{36.1} & \textbf{20.5} & \textbf{71.6} \up{3.7} & \textbf{55.1} & {\ul 64.9} & \textbf{89.9} & {\ul 90.4} & \textbf{74.8} \up{8.2} \\
\bottomrule
\end{tabular}%
}
\caption{Experimental results on four open-domain multi-hop reasoning datasets: HotpotQA, MuSiQue, 2WikiMQA and Bamboogle. Best and second results are highlighted by \textbf{bold} and {\ul underline}. 
The evaluation metric is F1.
All experiments are done with \textit{gpt-3.5-turbo-instruct} through in-context learning.
All baselines are instantiated with both sparse retriever and search engine.
$\diamondsuit:$ One-time retrieval.
$\spadesuit:$ Iterative retrieval.
$\clubsuit:$ Sub-question retrieval.
}
\label{tab:main_results}
\end{table*}

\subsection{Implementation Details}
\label{subsec:details}
In the main experiment, we use GPT-3.5-turbo as the backbone. Since OpenAI has deprecated GPT-3.5 (text-davinci series), we reproduce the baselines with GPT-3.5-turbo for a fair comparison\footnote{\url{https://platform.openai.com/docs/deprecations}}.
Please refer to Appendix~\ref{appendix:details} for more details.

\subsection{Baselines}
\label{subsec:baselines}
\paragraph{Standard Prompting~\citep{fewshot-prompting}}
directly generates the final answer with few-shot demonstrations. We use a 20-shot demonstration.

\paragraph{Chain-of-Thought Prompting~\citep{fewshot-cot}}
generates reasoning steps before the final answer. We use a 20-shot demonstration.

\paragraph{One-time Retrieval}
involves using the original question as the query, concatenating sparse retrieval and Google search results into the prompt to guide the model to perform CoT reasoning.

\paragraph{Self-Ask~\citep{bamboogle-selfask}}
adopts an iterative approach to decompose complex questions. It generates sub-questions iteratively based on existing reasoning, and then retrievals and answers them until reaching the final answer.

\paragraph{IRCoT~\citep{mhqa-ircot}}
interleaves retrieval-augmented reasoning and reasoning-augmented retrieval until the retrieval information is sufficient to answer the question.

\paragraph{FLARE~\citep{mhqa-flare}}
dynamically adjusts the retrieval timing based on the confidence of reasoning and conducts retrieval based on upcoming reasoning sentences.
\paragraph{ProbTree~\citep{mhqa-ptot}}
parses the question into a tree, employing logprobs-based sub-question aggregation to obtain the final answer.

\section{Experimental Results}
\label{sec:results}
\subsection{Main Results}
\label{subsec:main_results}
The experimental results on four multi-hop reasoning datasets are presented in Table~\ref{tab:main_results}.
We observe that one-time retrieval can improve performance in HotpotQA, 2WikiMQA and Bamboogle.
However, when facing the more challenging MuSiQue, one-time retrieval even impairs performance (24.7 $\to$ 16.4),
which indicates that inaccurate retrieval and knowledge conflicts may exacerbate hallucination.

The iterative retrieval approach has made notable progress compared to one-time retrieval by offering more comprehensive knowledge.
Additionally, methods based on sub-question retrieval have more explicit queries, which leads to a further improvement in retrieval accuracy, serving as the best-performing baseline method.

As shown in Table~\ref{tab:main_results}, our method demonstrates superiority over all baselines.
It outperforms the previous state-of-the-art, ProbTree, on all four datasets: HotpotQA (+2.5), MuSiQue (+4.0), 2WikiMQA (+3.7), and Bamboogle (+8.2).
We attribute the improvement to three aspects:
(i) Question decomposition leads to more accurate retrieval.
Compared to iterative retrieval based on generation content, sub-questions serve as clearer retrieval queries.
(ii) Complementary reasoning facilitates collaboration between diverse knowledge sources.
Our method dynamically selects and leverages knowledge from multiple sources with fine granularity, thereby mitigating knowledge conflicts and omissions.
(iii) Probability-based beam aggregation mitigates cascading errors and optimizes the reasoning trajectories through consistency.

\begin{figure}[t]
    \centering
    \includegraphics[clip, width=\linewidth]{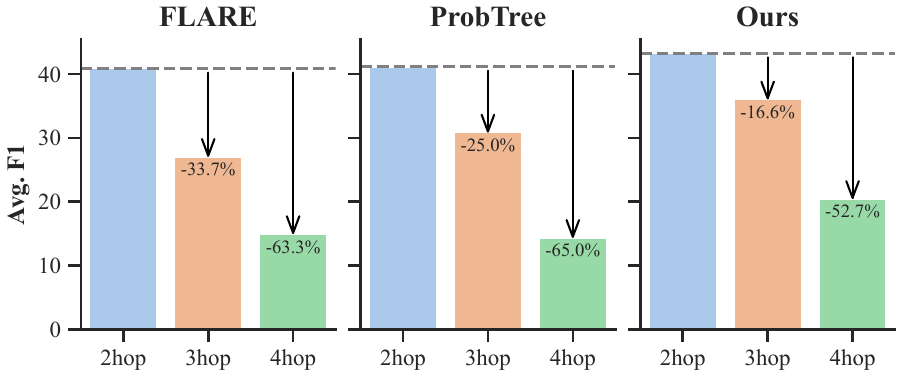}
    \caption{Performance gap with different reasoning steps. 
    We adopt the original split in MuSiQue and report the average f1 score in each subset.
    As the number of reasoning steps escalates, the model's performance declines.
    Our method exhibits a slower performance decline as reasoning steps increase, indicating its ability to effectively alleviate cascading errors.}
    \label{fig:234hop}
\end{figure}

We observe significant improvements, particularly for \textit{Comp} questions.
Comparison questions involve interactions among sub-questions.
With the aid of beam aggregation, the model can enumerate various combinations during reasoning, thereby encompassing a broader spectrum of possibilities (65.9 $\to$ 74.2, 81.7 $\to$ 89.9).
Additionally, our method is proficient at addressing complex questions (3/4-hop), as shown in Figure~\ref{fig:234hop}.
\label{subsec:exp_hop}
Our method exhibits a slower performance decline as reasoning depth increases (8.4\% in 3-hop and 12.3\% in 4-hop),
indicating its ability to precisely aggregate answers and effectively alleviate cascading errors.

\subsection{Results on Open-source Model}
\label{subsec:generalization_exp}
\begin{table}[t]
\centering
\setlength\tabcolsep{4pt}
\resizebox{\linewidth}{!}{%
\begin{tabular}{lcccc}
\toprule
\textbf{Methods} & \textbf{HotpotQA} & \textbf{MuSiQue} & \textbf{2WikiMQA} & \textbf{Bamboogle} \\
\midrule
\multicolumn{5}{l}{\textbf{\textit{Close-book Reasoning}}} \\
SP & 23.1 & 4.3 & 8.5 & 11.0 \\
CoT & 31.3 & 16.1 & 34.2 & 33.0 \\
\midrule
\multicolumn{5}{l}{\textbf{\textit{Retrieval-augmented Reasoning}}} \\
OneR & 43.6 & 11.4 & 36.3 & 28.1 \\
FLARE & 45.7 & 20.1 & 39.6 & 41.3 \\
ProbTree & 50.4 & 27.0 & 59.9 & 61.1 \\
\textbf{Ours} & \textbf{55.2} & \textbf{32.3} & \textbf{63.2} & \textbf{74.0}  \\
\midrule
\rowcolor{Gray}
Ours (GPT-3.5) & 62.9 & 36.9 & 71.6 & 74.8 \\
\bottomrule
\end{tabular}
}
\caption{
Experimental results on \textit{Mistral-7B}.
Our method still outperforms all baselines and is compatible with \textit{GPT-3.5-turbo}, suggesting its generalizability.
}
\label{tab:slm_results}
\end{table}

To prove the generalizability of our method to various models, we also conduct experiments on open-source models.
We select Mistral-7B~\citep{mistral}, the state-of-the-art LLM among similar scales.
As shown in Table~\ref{tab:slm_results}, \ours{} significantly outperforms previous SOTA on all four datasets, demonstrating its model-agnostic nature and effectiveness.
Furthermore, it is comparable to GPT-3.5-turbo on the Bamboogle dataset.

\subsection{Ablation Study}
\label{subsec:ablation}

\paragraph{Effect of Multi-source Reasoning}
We conduct two sets of ablations: removing a single knowledge source and removing a type of knowledge, as shown in Table~\ref{tab:ablation}(a).
Removing any knowledge source results in a certain performance decrease, suggesting that each type of knowledge contributes to reasoning (a.1-a.4). 
Furthermore, the declining trends (4.1\% internal knowledge and 9.7\% external knowledge) indicate that external knowledge makes a greater contribution compared to internal knowledge.
Using only internal knowledge for reasoning results in a severe performance drop (a.5).
Nevertheless, our method still outperforms chain-of-thought reasoning, which reflects the effectiveness of aggregation reasoning.
Completely removing internal knowledge leads to a substantial decline, as shown in (a.6).
This suggests that internal knowledge can complement external retrieval, proving the effectiveness of our method in knowledge collaboration.

\begin{table}[t]
\centering
\setlength\tabcolsep{4pt}
\resizebox{\linewidth}{!}{%
\begin{tabular}{lcc}
\toprule
\textbf{Setting} & \textbf{2WikiMQA} & \textbf{MuSiQue} \\
\midrule
\oursbf{} & \textbf{71.6}  & \textbf{36.9} \\
\midrule
\multicolumn{3}{l}{\textit{(a) Multi-Source Knowledge}} \\
~~1. w/o closebook & 69.2 & 35.1 \\
~~2. w/o parameter & 68.3 & 35.6 \\
~~3. w/o wikipedia & 65.8 & 33.1 \\
~~4. w/o web search & 63.4 & 32.7 \\
~~5. internal only & 52.0 & 27.3 \\
~~6. external only & 68.5 & 33.4 \\
\midrule
\multicolumn{3}{l}{\textit{(b) Beam Aggregation}} \\
~~1. polarization aggregation & 65.9 & 30.9 \\
~~2. w/o probability & 67.8 & 35.4 \\
~~3. greedy aggregation & 70.1 & 36.2 \\

\bottomrule
\end{tabular}%
}
\caption{Ablation results on multi-source knowledge reasoning and beam aggregation. 
Internal only: conduct internal reasoning only. 
External only: conduct external reasoning only.
W/o probability: do not distinguish the weights of aggregated answers.
Polarization aggregation: aggregate answers with logprobs.
}
\label{tab:ablation}
\end{table}

\paragraph{Effect of Beam Aggregation}
To validate the effectiveness of beam aggregation, we first employ log-prob-based aggregation, which can only select a single knowledge source.
As shown in Table~\ref{tab:ablation}(b), this leads to a notable decline, suggesting that coarse-grained polarized aggregation struggles to effectively integrate knowledge, thus underscoring the superiority of our beam aggregation.
Probabilistic aggregation enables the distinction of the importance of answers.
To investigate its effect, we conduct an ablation where the aggregated answers are treated with equal weights, as shown in (b.2).
The performance drops by 5\% and 4\% on two datasets, suggesting that prioritizing reasoning trajectories can elicit better reasoning.
Finally, we investigate the effect of candidate size.
Larger candidate sizes result in broader reasoning breadth, but they also increase reasoning overhead.
When it is set to 1, greedy aggregation strategy is employed.
It is a cost-efficient variant of beam aggregation reasoning.
As shown in (b.3), it causes a slight performance decrease, indicating that some erroneous reasoning may be corrected as the reasoning process progresses with the help of a broader scope of reasoning.
We conduct a detailed analysis of reasoning performance and overheads in section~\ref{subsec:pareto}.

\section{Analysis}
\label{sec:analysis}

\begin{figure}[t]
    \centering
    \includegraphics[clip, width=\linewidth]{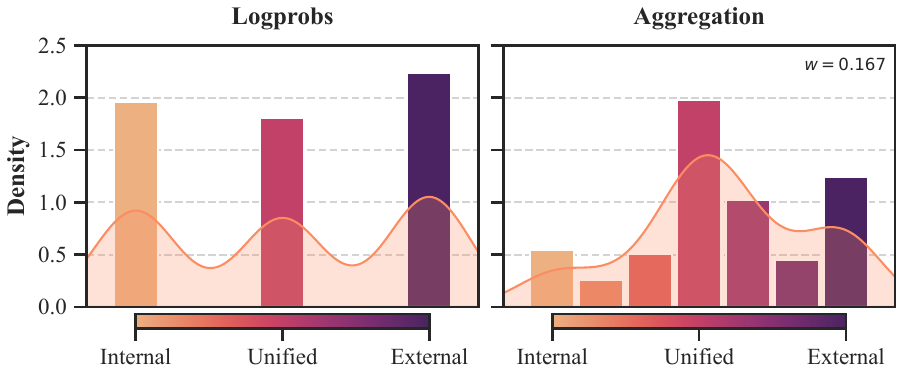}
    \caption{Distribution of knowledge integration in reasoning.
\textbf{Unified} represents the integration of multi-source knowledge in reasoning, while the two ends represent reliance on single-source knowledge.
The bars represent original discrete distributions, and the curve is the kernel density estimate (KDE).
    }
    \label{fig:preliminary}
\end{figure}
\begin{figure}[t]
    \centering
    \includegraphics[clip, width=\linewidth]{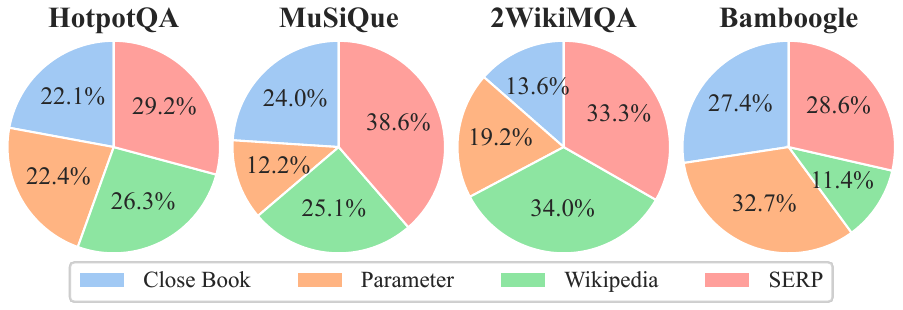}
    \caption{The contribution of each reasoning strategy to the final answer in percentage.
    HotpotQA is balanced, while MuSiQue and 2WikiMQA leans to external knowledge, and Bamboogle favors internal knowledge.}
    \label{fig:percentage}
\end{figure}
\begin{figure}[t]
    \centering
    \includegraphics[clip, width=\linewidth]{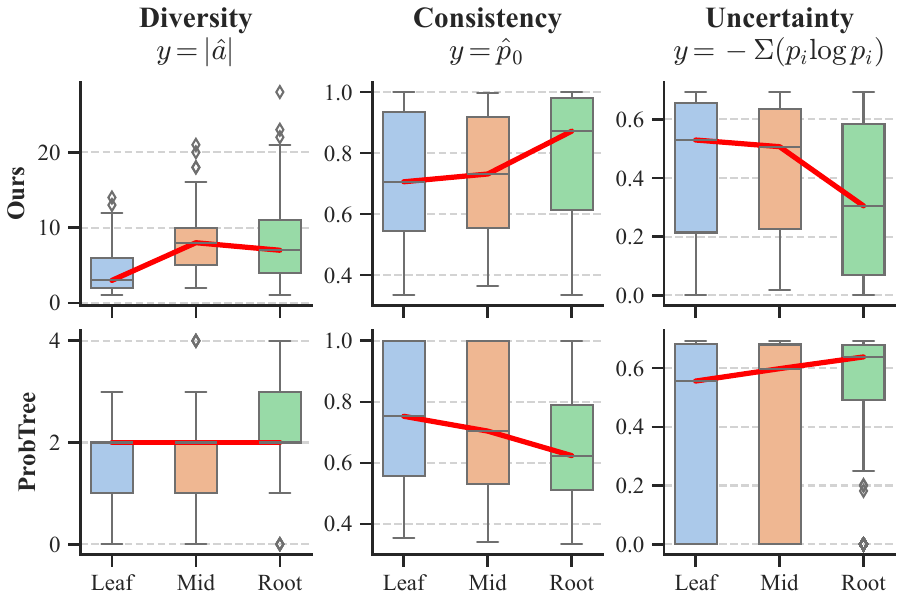}
    \caption{The tendency of candidate answers distribution from leaf node to root.
We define three metrics to measure the distribution.
Diversity: The number of distinct answers.
Consistency: The proportion of the majority answer.
Uncertainty: The information entropy of answer distribution.
}
    \label{fig:trend}
\end{figure}
\begin{figure}[t]
    \centering
    \includegraphics[clip, width=\linewidth]{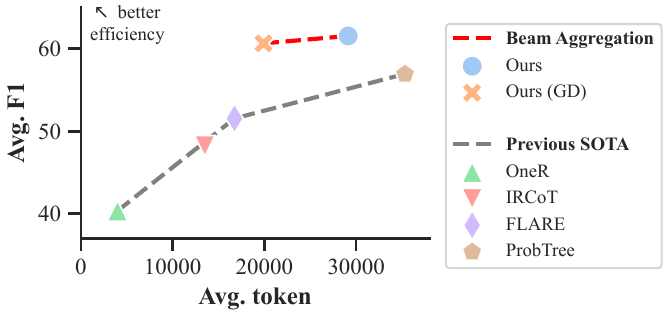}
    \caption{Pareto chart for token consumption and performance (F1). 
The upper-left quadrant indicates higher efficiency.
Results are averaged over four datasets.}
    \label{fig:pareto}
\end{figure}

\subsection{BeamAggR Facilitates Knowledge Collaboration}
To investigate the impact of \ours{} on knowledge collaboration, we conduct a preliminary experiment.
We track the model's utilization of knowledge from different sources during the reasoning process, as depicted in Figure~\ref{fig:preliminary}.
The polarity aggregation method based on log-probs excessively relies on reasoning on a single source of knowledge.
In more than 2/3 of cases, it can only utilize knowledge from a single source.
In contrast, our approach facilitates better knowledge collaboration in reasoning.
In summary, our beam aggregation can effectively employ knowledge from multiple sources at a finer granularity.

Furthermore, we carry out a more detailed examination of the proportions of each type of knowledge in reasoning.
We track the contribution of different knowledge sources to the final answer, as illustrated in Figure~\ref{fig:percentage}.
Disparities in reasoning contributions across various datasets are observed. 
2WikiMQA and MuSiQue tend to favor external reasoning, Bamboogle leans towards internal reasoning, and HotpotQA strikes a balance among different types of knowledge.
It is noteworthy that, without manual adjustment of the weight of knowledge, \ours{} can adaptively adjust its proportions during the reasoning process. 
This indicates that our method can effectively integrate multi-source knowledge.

\subsection{Trends of Answer Aggregation in Bottom-up Reasoning}
We conduct a detailed analysis of the trends in answer aggregation throughout the reasoning process.
To assess the characteristics of the answer distribution during the bottom-up reasoning process, we define three metrics: diversity, consistency, and uncertainty.
The definition and tendency are shown in Figure~\ref{fig:trend}.
It can be observed that through the reasoning process, the diversity of answers improves, indicating its exploration of a wider range of possibilities.
In contrast, consistency and uncertainty continue to decrease as the reasoning depth increases.
This suggests that our method is capable of filtering out reasoning from erroneous during the bottom-up aggregation, dynamically choosing appropriate knowledge sources and reasoning trajectories.
Conversely, methods based on log-probs aggregation are affected by inaccurate aggregation and cascading errors, thus unable to achieve these, highlighting the superiority of our method.

\subsection{Analysis of Reasoning Cost}
\label{subsec:pareto}

Retrieval-augmented generation often involves frequent invocation of LLMs, resulting in significant computational overhead.
We compare the performance and overhead of five RAG methods.
As illustrated in Figure~\ref{fig:pareto}, previous methods exacerbate reasoning overhead while improving performance.
In contrast, our method not only surpasses the previous SOTA in performance but also incurs lower overhead.
Moreover, our method can further reduce reasoning overhead through greedy aggregation.
In summary, \ours{} is Pareto efficient in balancing performance and reasoning overhead.
Detail statistics can be found in Table~\ref{tab:token_consume}.

\section{Conclusion}
\label{sec:conclusion}

This paper introduces BeamAggR for knowledge-intensive multi-hop reasoning.
BeamAggR utilizes a divide-and-conquer strategy, breaking down complex questions into a tree structure and conducting reasoning in a bottom-up fashion.
At each step of reasoning, BeamAggR builds upon previous candidates to identify the most likely reasoning path.
Furthermore, it incorporates multi-source reasoning to enhance knowledge collaboration.
Overall, BeamAggR facilitates multi-hop reasoning through precise sub-question retrieval, efficient knowledge collaboration, accurate answer aggregation, and broad exploration of reasoning trajectories.
Extensive experiments on four open-domain multi-hop reasoning datasets demonstrate its effectiveness.

\section*{Limitations}
\label{sec:limitations}

Our method involves beam combination across multiple candidates and self-consistency in each reasoning step, thereby increasing reasoning overhead. 
This overhead can be mitigated through greedy aggregation reasoning.
The effectiveness of our framework is contingent upon the accurate decomposition of questions, which may pose significant challenges for large language models.
Currently, we only use internal knowledge and unstructured external knowledge for reasoning.
In future work, structured external knowledge, such as knowledge bases, can be integrated into reasoning to provide more comprehensive knowledge repositories~\citep{cok}.

\section*{Acknowledgements}
The research in this article is supported by the National Key Research and Development Project (2021YFF0901602), the National Science Foundation of China (U22B2059, 62276083), and Shenzhen Foundational Research Funding (JCYJ20200109113441941), Major Key Project of PCL (PCL2021A06).
Ming Liu and Qianglong Chen are the corresponding authors.

\nocite{missref-mhqa1,missref-mhqa2}

\bibliography{custom}

\appendix
\section{Appendix}
\label{sec:appendix}

\subsection{Task Defination}
\label{appendix:task_def}
This paper focuses on open-domain multi-hop question-answering tasks.
The task's input is a multi-hop question $Q$ that requires multi-step reasoning to solve, with each step of reasoning necessitating specific knowledge.
Given $Q$ as the query, the retriever $\mathcal{R}$ retrieves relative documents $\mathcal{D}=\{d_i\}_{i=1}^{|\mathcal{D}|}$.
The retrieved documents $\mathcal{D}$ are then fed as context into the model for knowledge reasoning.
$y=\mathrm{LM}(Q, \mathcal{D})$.
Our method employs LLMs with few-shot prompts and external retrieval to address the tasks.

\subsection{Implementation Details of BeamAggR }
\label{appendix:details}
\paragraph{Retrieval Setup}
We use the October 2017 Wikipedia dumps\footnote{\url{https://hotpotqa.github.io/wiki-readme.html}} as the retrieval corpus, employing BM25~\citep{bm25} implemented by Elasticseach as the sparse retriever. 
For the web search, we use the Google SERP provided by Serper\footnote{\url{https://serper.dev/}}.
For SERP results, we consider the top-3 organic results as well as the answer box (if available), resulting in 3 or 4 retrieval documents.
We use wikipedia\footnote{\url{https://pypi.org/project/wikipedia/}} package to access raw Wikipedia content,
and perform fuzzy matching to extract snippets in the document with fuzzywuzzy\footnote{\url{https://pypi.org/project/fuzzywuzzy/}} package.
To ensure experimental fairness, we incorporate Google search results as additional knowledge for the baseline methods.

\paragraph{Hyperparameters}
Our main experiments are conducted using \textit{gpt-3.5-turbo-instruct} provided by Azure OpenAI 12-01-preview version.
The experiments on open-source models are conducted using Mistral-7B~\citep{mistral}.
Table~\ref{tab:hyperparameters} shows detailed hyperparameters.

\paragraph{Evaluation}
We use token-level F1 as evaluation metric. 
Besides, if entity aliases are available, we calculate the maximum F1 score between the prediction and each ground truth alias.

\paragraph{Prompts}
For datasets except for Bamboogle, we adopt the question decomposition results provided by \citet{mhqa-ptot}.
We convert their decomposition into our format and filter out those with invalid decomposition formats, regenerating them as necessary.
For open-source models, limited by their capabilities, they conduct reasoning based on the question decomposition provided by GPT-3.5.

\begin{table}[h]
\centering
\setlength\tabcolsep{6pt}
\resizebox{0.85 \linewidth}{!}{%
\begin{tabular}{lc}
\toprule
\bf Hyperparameters &\bf Values\\

\midrule

\# Retrieval doc & 5 \\
\# Retieval serp & 3 or 4 \\
\# Parametric knowledge & 1 \\
\midrule
\# Closebook prompt & 24 \\
\# Parametric prompt & 5 \\
\# Wikipedia prompt & 3 \\
\# SERP prompt & 5 \\
\# Question Decomp. prompt & 24 \\
\midrule
Beam size & 2 \\
Temperature & 3 \\
Self-consistency ($\tau$ = 0.7) & 5  \\

\bottomrule
\end{tabular}
}
\caption{Hyperparameters of BeamAggR.}
\label{tab:hyperparameters}
\end{table}

\subsection{Details of Dataset}
\label{appendix:dataset}
We provide statistics of the datasets, along with examples of each question type, as shown in Table~\ref{tab:dataset}.
\paragraph{HotpotQA} consists of 2-hop Bridge and Comparison questions.
Bridge: Inferring through the bridge entity to complete the 2-hop question.
Comparison: Comparing two entities from the same category, including yes/no questions.

\paragraph{MuSiQue}
composites one-hop questions through bridge and composition into multi-hop complex questions, covering six categories of 2 to 4-hop questions.
Bridge: Composite two single-hop questions into a 2-hop question through a bridge entity.
Composition: A question is connected to two sub-questions via two bridge entities, constituting a 3-hop question.
Bridge and Composition: Combining 2-hop bridge questions and 3-hop composition questions in different order yields 4hop composition/bridge questions.

\paragraph{2WikiMQA} includes four types of questions: bridge, inference, comparison, and bridge comparison.
Bridge and comparison questions are similar to those in HotpotQA.
Inference questions are similar to bridge questions, except they use inference relationships instead of bridge entities, such as \textit{grandfather of}.
Bridge-comparison: Comparing entities from two bridge sub-questions.

\paragraph{Bamboogle} 
consists of 2-hop bridge questions, which are similar to 2hop (bridge) questions in MuSiQue.
All questions in Bamboogle cannot be directly answered through search engines.

\subsection{Details of Analysis}

\paragraph{Distribution of knowledge integration}
In order to facilitate a fair comparison between log-probs and probabilistic aggregation in the preliminary study, we limit our statistical analysis to the proportion of internal and external knowledge contained within the top-1 answers in atomic questions.
This ensures that the outcome of the statistics is not affected by the types of decomposition (iterative or sub-question retrieval) or different reasoning paths.
The experimental results are averaged from 500 entries each from the HotpotQA and Musique datasets.
We plot the discrete distributions (Histplot) and the kernel density estimate (KDEplot) in Figure~\ref{fig:preliminary}.
The y-axis is probabilistic density, for discrete distributions, the density is the percentage divided by the width of the bar. In this figure, the bar width $w$ is set to $1/6 \approx 0.167$.

\paragraph{Contribution of each knowledge source}
Beginning with the final answer (the top-1 answer from the root node), we trace the reasoning path utilized from top to bottom, counting the knowledge sources that participated in the voting for sub-question answers along this path. 
Finally, we compute the percentage of knowledge sources within the entire path.
To ascertain that this combination of knowledge aggregation led to the correct answer, we omit erroneous samples from our analysis.

\paragraph{Tendency of candidate answers distribution}
In the reasoning process, errors can be accumulated, leading to cascading errors.
Through multiple reasoning paths and probabilistic aggregation, \ours{} can gradually reduce the uncertainty.
To illustrate this, we classify reasoning hop into three types: leaf, mid, and root.
We statistically analyze the candidate answer distribution in each hop and use diversity, consistency, and uncertainty to measure this distribution.
Figure~\ref{fig:trend} shows a boxplot figure of these statistics.
We also line the median value of each hop to show the trend more clearly.
The analysis is conducted on the MuSiQue dataset, which has a clearer reasoning hop structure.

\begin{table}[t]
\centering
\setlength\tabcolsep{6pt}
\resizebox{\linewidth}{!}{%
\begin{tabular}{lcccccccc}
\toprule
\multirow{2.5}{*}{\textbf{Methods}} & \multicolumn{2}{c}{\textbf{HotpotQA}} & \multicolumn{2}{c}{\textbf{MuSiQue}} & \multicolumn{2}{c}{\textbf{2WikiMQA}} & \multicolumn{2}{c}{\textbf{Bamboogle}}  \\
\cmidrule(lr){2-3}  \cmidrule(lr){4-5} \cmidrule(lr){6-7} \cmidrule(lr){8-9} 
 & \textit{\#token} & \textit{f1} & \textit{\#token} & \textit{f1} & \textit{\#token} & \textit{f1} & \textit{\#token} & \textit{f1} \\
\midrule
\multicolumn{9}{l}{\textit{\textbf{One-Time Retrieval}}} \\
OneR & 4053 & 55.3 & 3941 & 16.4 & 3892 & 42.9 & 4082 & 46.8 \\
\midrule
\multicolumn{9}{l}{\textit{\textbf{Iterative Retrieval}}} \\
IRCoT & 13550 & 56.2 & 16877 & 24.9 & 13273 & 56.8 & 10347 & 55.0 \\
FLARE & 17793 & 56.1 & 19180 & 31.9 & 16651 & 60.1 & 13358 & 58.1 \\
\midrule
\multicolumn{9}{l}{{\textit{\textbf{Sub-question Retrieval}}}} \\
ProbTree & 25607 & 60.4 & 46431 & 32.9 & 39249 & 67.9 & 30004 & 66.6 \\
\textbf{Ours} & 23720 & 62.9 & 36336 & 36.9 & 30178 & 71.6 & 26276 & 74.8 \\
Ours (GA) & 17887 & 62.3 & 22522 & 36.2 & 21703 & 70.1 & 17507 & 74.0 \\
\bottomrule
\end{tabular}%
}
\caption{Detailed token cost per instance and performance in four datasets. 
GA: greedy aggregation.
(\S\ref{subsec:pareto})}
\label{tab:token_consume}
\end{table}

\paragraph{Token consumption}
We compare the efficiency between one-time retrieval (OneR), iterative retrieval (IRCoT, FLARE), and sub-question retrieval (ProbTree, Ours).
We use a Pareto chart to visualize the token consumption of each method, as shown in Figure~\ref{fig:pareto}.
The token consumption per instance and performance are averaged in four datasets.
The upper left indicates better efficiency (less token consumption and higher performance).
Detailed token consumption in each dataset is shown in Table~\ref{tab:token_consume}.
We will detail the method for calculating token consumption in the following section.

We measure the computational cost by evaluating the average token consumption per question.
Specifically, the cost of each instance includes prompt tokens (demonstrations, question, retrieved documents) and completion tokens (reasoning traces, answer).
For retrieval-augmented reasoning, the cost of a single API call is approximately 4000 tokens, whereas for non-retrieval reasoning, the cost is less than 1000 tokens.

\subsection{Prompts}
We provide manually annotated demonstrations on the 2WikiMQA dataset for reference.
Our prompts are derived from IRCoT\footnote{\url{https://github.com/StonyBrookNLP/ircot}}~\citep{mhqa-ircot} and ProbTree\footnote{\url{https://github.com/THU-KEG/ProbTree}}~\citep{mhqa-ptot}, to which we have made some modifications.
The question decomposition is carried out using 24-shot demonstrations (Figure~\ref{demo:question-decomp}).
For implicit internal knowledge, closed-book reasoning is conducted using 20-shot chain-of-thought demonstrations (Figure~\ref{demo:cb-reason}).
For explicit internal knowledge, we first have the LLM generate parametric knowledge (Figure~\ref{demo:knowledge-gen}), followed by knowledge reasoning using 5-shot demonstrations (Figure~\ref{demo:reason-param}).
For external knowledge reasoning based on web search and Wikipedia, we use prompts of 5-shot and 3-shot (Figure~\ref{demo:reason_serp},~\ref{demo:reason_doc}).

\subsection{Formal Definition of Notations}
We describe the formal definition of the notations used in the algorithm pseudocode and formulas in this paper, as shown in Table~\ref{tab:symbol}.

\begin{table*}[h]
\centering
\begin{tabularx}{\linewidth}{lcX}
\toprule
Question Type & \#Examples &Question Example \\
\midrule
\multicolumn{2}{l}{\textit{\textbf{(a) HotpotQA}}} \\
Bridge &  412 & Which team does the player named 2015 Diamond Head Classic’s MVP play for? \\
Comparison &  88 & Did LostAlone and Guster have the same number of members? \\

\midrule
\multicolumn{2}{l}{\textit{\textbf{(b) MuSiQue}}} \\
2hop (Bridge) & 254 & Who succeeded the first President of Namibia? \\
3hop1 (Bridge) & 122 & What currency is used where Billy Giles died? \\
3hop2 (Composition) & 32 & When was the first establishment that McDonaldization is named after, open in the country Horndean is located? \\
4hop1 (Bridge) & 51 & When did Napoleon occupy the city where the mother of the woman who brought Louis XVI style to the court died?\\
4hop2 (Bridge + Composition) & 16 & How many Germans live in the colonial holding in Aruba’s continent that was governed by Prazeres’s country?\\
4hop3 (Composition + Bridge) & 25 & When did the people who captured Malakoff come to the region where Philipsburg is located?\\

\midrule
\multicolumn{2}{l}{\textit{\textbf{(c) 2WikiMQA}}} \\
Comparison & 119 & Who was born first, Albert Einstein or Abraham Lincoln?\\
Inference & 79 & Who is the maternal grandfather of Abraham Lincoln?\\
Bridge & 197 & Who is the founder of the company that distributed La La Land film?\\
Bridge-comparison & 105 & Which movie has the director born first, La La Land or Tenet?\\

\midrule
\multicolumn{3}{l}{\textit{\textbf{(d) Bamboogle}}} \\
2hop (Bridge) & 125 & When did the last king from Britain’s House of Hanover die? \\

\bottomrule
\end{tabularx}
\caption{Statistics and examples of each question type in HotpotQA, MuSiQue, 2WikiMQA and Bamboogle.
It should be noted that some instances in these datasets trigger the content filter of  Azure OpenAI API, where the model refuses to respond. 
And thus we have omitted these filtered samples when calculating metrics.
}
\label{tab:dataset}
\end{table*}

\clearpage

\begin{table*}[h]
\centering
\begin{tabularx}{\linewidth}{ccX}
\toprule
Symbol & Dim / Contains & Description \\
\midrule
\multicolumn{3}{l}{\textit{\textbf{(a) Inputs}}} \\
$Q$ & - & Complex knowledge-intensive multi-hop question. \\
$R(\dots)$ & - & Multi-source knowledge retriever. \\
$LLM(\dots)$ & - & Large language model prompted with few-shot demonstration. $LLM(\dots,R)$ indicates that the reasoning is augmented by the retriever. \\
$k$ & - & Hyper-parameter: Candidate size in beam aggregation. \\
\midrule
\multicolumn{3}{l}{\textit{\textbf{(b) Question Tree}}} \\
$Q_{decomp}$ & $\{N^{(1)}, N^{(2)} \dots\}$ & Decomposed questions tree, contains tree nodes. The root node $N^{\textrm{root}}$ represents the original question, and the children of a node $\textrm{sons}(N^{(i)})$ are sub-questions. \\
$N^{(i)}$ & $\{q^{(i)}, ~\va^{(i)}, ~\vp^{(i)}\}$ & Tree node $i$, contains the sub-question, answers and probabilities. \\
$q^{(i)}$ & $\R$ & Sub-question for node $i$, obtained in decompose step. The question may contain placeholder tokens (e.g. \#1) which need to be replaced by the answer in sub-questions (sons) during reasoning. \\
$\va^{(i)}, ~\vp^{(i)}$ & $\R^{k}, ~\R^{k}$ & Top-k candidate answers and corresponding probabilities in node $i$, obtained in bottom-up reasoning step. \\
\midrule
\multicolumn{3}{l}{\textit{\textbf{(c) Variables in Reasoning Step}}} \\
$~\hat{\va}, ~\hat{\vp}$ & $\R^{?}, ~\R^{?}$ & Intermediate results for answers and corresponding probabilities, need to be further voted or aggregated. \\
$\va^t, ~\vp^t$ & $\R^{?}, ~\R^{?}$ & Same as above. \\
$\vc', ~p'$ & $\R^{\vert\textrm{sons}\vert}, \R$ & One combination of sub-questions answer and corresponding probabilities. A combination is one element in the cartesian product of the children's answers $\vc' \in \bigtimes\limits_{i \in \textrm{sons}} \va^{(i)} = \va^{\textrm{sons}_1} \times \va^{\textrm{sons}_2} \dots$ and $p'$ is the production of children's probabilities for this combination. \\
\midrule
\multicolumn{3}{l}{\textit{\textbf{(d) Functions in Pseudocode}}} \\
$\textrm{PostOrderTraverse}(Q)$ & $[N] \rightarrow [N]$ & Return a new node sequence in post-order (child0 < child1 < ... < root). \\
$\textrm{CartProd}(sons)$ & $[N] \rightarrow [[a], ~p]$ & Cartesian product of the answers and the associated probabilities of the children. For instance, if a node has two children $x, y$, this function will return a set $\{ \langle a^{(x)}_i, a^{(y)}_j \rangle, p^{(x)}_i p^{(y)}_j ~|~ i, j = 1, 2, \dots, k \}$. \\
$\textrm{MaskFill}(q, c)$ & $q, [a] \rightarrow q$ & Replaces the placeholder token ($\#i$) in the sub-question $q$ with candidate answer $c_i$. \\
$\textrm{Vote}(\va)$ & $[a] \rightarrow [a], [p]$ & Deduplicate answers and obtain a probabilities distribution based on their frequency. Returns are arranged in descending order of probabilities. \\
$\textrm{Aggr}(\va, \vp)$ & $[a], [p] \rightarrow [a], [p]$ & Return the unique answers and their normalized probabilities, arranged in descending order of probabilities. \\
\bottomrule
\end{tabularx}
\caption{The formal definition of notations used in the algorithms and formulas within this paper.}
\label{tab:symbol}
\end{table*}

\clearpage

\begin{figure*}[h]
    \centering
    \examplefigure{Reasoning Example on \textbf{Bamboogle}}{
\textbf{Question}: The fourth largest city in Germany was originally called what? \\
\textbf{Decomposition}: \\Q1. What is the fourth largest city in Germany? \\Q2. What was \#1 originally called? \newline

\textbf{Q1}: What is the fourth largest city in Germany? \\
closebook answer: [Frankfurt, 3], [Cologne, 2] \\
parametric answer: [Cologne, 5] \\
document answer: [Regensburg, 3] \\
serp answer: [Darmstadt, 5] \\
> aggregated answer: [Cologne, 0.6607], [Darmstadt, 0.3392] \newline

\textbf{Q2-1}: What was [Cologne] originally called?  \\
\textbf{Q2-2}: What was [Darmstadt] originally called? \\
> aggregated answer: [Colonia Claudia Ara Agrippinensium, 0.5229], [Colonia Agrippina,
          0.1378], [Darmundestat, 0.2988], [the Grand Duchy of Hesse, 0.0404] \newline

\textbf{Q}: The fourth largest city in Germany was originally called what? \\
aggregated answer: [Colonia Claudia Ara Agrippinensium, 0.6363], [Darmundestat, 0.3636] \\
> final answer: Colonia Claudia Ara Agrippinensium \textcolor{red}{\checkmark} \\
ground truth: Colonia Claudia Ara Agrippinensium

}
    \caption{An example of reasoning on Bamboogle dataset}
    \label{example:bamboogle}
\end{figure*}
\begin{figure*}[h]
    \centering
    \examplefigure{Question Decomposition Examples}{

    \textit{(a) MuSiQue}\\
    \textbf{Question}: Who played who sang is she really going out with him in the who influenced Beyonce movie?\\
    \textbf{Decomposition}: \\Q1. Which movie was influenced by the Who? \\Q2. Who sang 'Is She Really Going Out With Him'? \\Q3. Who played \#2 in \#1? \newline

    \textit{(b) HotpotQA}\\
    \textbf{Question}: What German state was Karl Julius Perleb born in?\\
    \textbf{Decomposition}: \\Q1. Where was Karl Julius Perleb born? \\Q2. What German state is \#1 in? \newline

    \textit{(c) 2WikiMQA}\\
    \textbf{Question}: Which film has the director who was born later, Money On The Street or She-Devils On Wheels?\\
    \textbf{Decomposition}: \\Q1. Who is the director of film Money On The Street? \\Q2. When was \#1 born? \\Q3. Who is the director of film She-Devils On Wheels? \\Q4. When was \#3 born? \\Q5. Which film has the director who was born later, Money On The Street or She-Devils On Wheels? (\#2, \#4) \newline

    \textit{(d) Bamboogle}\\
    \textbf{Question}: Who built the fastest air-breathing manned aircraft? \\
    \textbf{Decomposition}: \\Q1. What is the name of the fastest air-breathing manned aircraft? \\Q2. Who built \#1?
   
}
    \caption{Examples of question decomposition}
    \label{example:decomposition}
\end{figure*}
\clearpage

\begin{figure*}[h]
    \centering
    \promptfigure{Demonstrations of \textbf{Question Decomposition}}{
Question: Who is the performer of Live at this studio that employs the person who coined the term theatre of the absurd?\\
Decompose: {"Who is the performer of Live at this studio that employs the person who coined the term theatre of the absurd?": ["Where did the person who coined the term the theatre of the absurd work?", "Who is the performer at the Live at the \#1 event?"], "Where did the person who coined the term the theatre of the absurd work?": ["Who coined the term the theatre of the absurd", "Where is \#1 worked?"]}\newline

Question: Who is the general treasurer of the state where Israel Arnold House is located?\\
Decompose: {"Who is the general treasurer of the state where Israel Arnold House is located?": ["What state is Israel Arnold House located?", "Who is the general treasurer of \#1?"]}\newline

Question: What weekly publication in the place of death of George Townsend is issued by the employer of the Yale labor historian who advised younger historians?\\
Decompose: {"What weekly publication in the place of death of George Townsend is issued by the employer of the Yale labor historian who advised younger historians?": ["Where the Yale labor historian who advised younger labor historians works?", "Where did George Townsend die?", "What weekly publication in \#2 is issued by \#1?"], "Where the Yale labor historian who advised younger labor historians works?": ["Which Yale labor historian advised other younger labor historians?", "Where \#1 works?"]}\newline

Question: When was the SEC championship game between the winner of the most national titles in NCAA football and Georgia?\\
Decompose: {"When was the SEC championship game between the winner of the most national titles in NCAA football and Georgia?": ["Who has the most national titles in NCAA football?", "When was the SEC championship game between \#1 and georgia?"]}\newline

Question: Who sings Never Say Never with the performer of As Long as You Love Me?\\
Decompose: {"Who sings Never Say Never with the performer of As Long as You Love Me?": ["Who is the performer of As Long as You Love Me?", "Who sings Never Say Never with \#1?"]}\newline
......}
    \caption{Demonstrations of question decomposition. (24 shot)}
    \label{demo:question-decomp}
\end{figure*}
\begin{figure*}[h]
    \centering
    \promptfigure{Demonstrations of \textbf{Close-book Reasoning}}{
Question: When did the director of film Hypocrite (Film) die?\\
Answer: The film Hypocrite was directed by Miguel Morayta. Miguel Morayta died on 19 June 2013. So the answer is **19 June 2013**.\newline

Question: Do director of film Coolie No. 1 (1995 Film) and director of film The Sensational Trial have the same nationality?\\
Answer: Coolie No. 1 (1995 film) was directed by David Dhawan. The Sensational Trial was directed by Karl Freund. David Dhawan's nationality is India. Karl Freund's nationality is Germany. Thus, they do not have the same nationality. So the answer is **no**.\newline

Question: Are both Kurram Garhi and Trojkrsti located in the same country?\\
Answer: Kurram Garhi is located in the country of Pakistan. Trojkrsti is located in the country of Republic of Macedonia. Thus, they are not in the same country. So the answer is **no**.\newline

Question: Who was born first out of Martin Hodge and Ivania Martinich?\\
Answer: Martin Hodge was born on 4 February 1959. Ivania Martinich was born on 25 July 1995. Thus, Martin Hodge was born first. So the answer is **Martin Hodge**.\newline

Question: Which album was released more recently, If I Have to Stand Alone or Answering Machine Music?\\
Answer: If I Have to Stand Alone was published in the year 1991. Answering Machine Music was released in the year 1999. Thus, of the two, the album to release more recently is Answering Machine Music. So the answer is **Answering Machine Music**.\newline
......}
    \caption{Demonstrations of close-book reasoning on 2WikiMQA dataset. (20 shot)}
    \label{demo:cb-reason}
\end{figure*}
\begin{figure*}[h]
    \centering
    \promptfigure{Instruction of \textbf{Parametric Knowledge Generation}}{
Provide the necessary background knowledge to answer the given question. \\
Question: \{\}\\
Knowledge: }
    \caption{Instruction of parametric knowledge generation. (zeroshot)}
    \label{demo:knowledge-gen}
\end{figure*}
\begin{figure*}[h]
    \centering
    \promptfigure{Demonstrations of \textbf{Explicit Parametric Reasoning}}{
Given a question and the relevant documents, answer the question and explain why. If you are unsure, answer Unknown.\newline

\#1 Document:\\
Kurram Garhi  Kurram Garhi is a town located in the Kurram District of Khyber Pakhtunkhwa, Pakistan. It is situated on the bank of Kurram River and is approximately 12 kilometers away from the city of Parachinar, the district's headquarters.  The word "Kurram" is derived from the Sanskrit word "Kramar," which means "a place to live." It is believed that Kurram Garhi was named by the Hindu king, Raja Karanpal, who ruled the area in the 10th century.  The town has a rich history and has been a significant strategic location throughout the centuries. It has been a part of various empires and has seen many battles between different rulers. In the 18th century, Kurram Garhi was under the control of the Mughal Empire, and later it became a part of the Durrani Empire.\\
Question: Which country is Kurram Garhi located in?\\
Answer: Kurram Garhi is located in the country of Pakistan. So the answer is **Pakistan**.\newline

\#1 Document:\\
Monte Galbiga is a mountain located in the province of Como, Lombardy, Italy. It has an elevation of 1,690 meters (5,545 feet) above sea level. The mountain is also known as the "Balcone d'Italia" (Balcony of Italy) due to its panoramic views of Lake Como and the surrounding mountains.  The name "Galbiga" is derived from the Lombard word "galb" which means "height". The mountain is a popular destination for hikers and offers various trails and viewpoints. It is also a popular spot for paragliding and hang gliding.  In addition to its natural beauty, Monte Galbiga also has historical significance. During World War II, it was used as a strategic observation point by the Italian army. Remains of fortifications and bunkers can still be found on the mountain. \\
Question: In which country is the mountain Monte Galbiga located?\\
Answer: The mountain Monte Galbiga is located in Italy. So the answer is **Italy**.\newline

......}
    \caption{Demonstrations of explicit parametric reasoning on 2WikiMQA dataset. (5 shot)}
    \label{demo:reason-param}
\end{figure*}
\begin{figure*}[h]
    \centering
    \promptfigure{Demonstrations of \textbf{Open-book Reasoning (Wikipedia)}}{
Given a question and the relevant Wikipedia text, answer the question and explain why. If you are unsure, answer Unknown.\newline

\#1 Wikipedia Title: Hypocrite (film)\\
Text: Hypocrite (Spanish: Hipócrita..!) is a 1949 Mexican thriller film directed by Miguel Morayta ...\\
\#2 Wikipedia Title: When the Legends Die\\
Text: When The Legends Die is a 1963 novel, by Hal Borland, and a DeLuxe Color film released ...\\
\#3 Wikipedia Title: Who Is the Guilty?\\
Text: Who is the Guilty? ( sometimes" Who is to Blame?") is a 1925 Georgian silent film ...\\
\#4 Wikipedia Title: Miguel Morayta\\
Text: Miguel Morayta( 15 August 1907 – 19 June 2013) was a Spanish film director and screenwriter ...\\
\#5 Wikipedia Title: Joselito vagabundo\\
Text: Joselito vagabundo(" Joselito Vagabond") is a 1966 Mexican film. It stars Sara García and is directed by ...\\
Question: When did the director of film Hypocrite (Film) die?\\
Answer: The film Hypocrite was directed by Miguel Morayta. Miguel Morayta died on 19 June 2013. So the answer is **19 June 2013**.\newline

......}
    \caption{Demonstrations of open-book reasoning (wikipedia) on 2WikiMQA dataset. (3 shot)}
    \label{demo:reason_doc}
\end{figure*}
\begin{figure*}[h]
    \centering
    \promptfigure{Demonstrations of \textbf{Open-book Reasoning (SERP)}}{
Please answer the question based on the snippet from Google search and provide an explanation. If you are unsure, answer Unknown.\newline

\#1 Wikipedia Title: Shah dynasty\\
Snippet: Dravya Shah was the youngest son of Yasho Brahma Shah, Raja (King) of Lamjung and grandson of Kulamandan Shah Khad, Raja (King) of Kaski ...\\
\#2 Wikipedia Title: List of monarchs of Nepal\\
Snippet: The monarchs of Nepal were members of the Shah dynasty who ruled over the Kingdom of Nepal from 1743 to its dissolution in 2008 ... \\
\#3 Wikipedia Title: Krishna Shah\\
Snippet:Krishna Shah (10 May 1938 – 13 October 2013) was an Indian-American/Gujarati film and theatre director, screenwriter, playwright, producer, and production/distribution executive ...\\
Question: Who is the child of Krishna Shah (Nepalese Royal)?\\
Answer: Krishna Shah was the father of Rudra Shah. So the answer is **Rudra Shah**.\newline

\#1 Wikipedia Title: Kurram Garhi Hydropower Plant\\
Snippet: Kurram Garhi Hydropower Plant (KGHPP) is a small, low-head, run-of-the-river hydroelectric power generation station of 4.0 megawatt generation capacity ...\\
\#2 Wikipedia Title: Kurram District\\
Snippet: Kurram District is a district in the Kohat Division of the Khyber Pakhtunkhwa province of Pakistan.The name Kurram comes from the river Kwarma ...\\
\#3 Wikipedia Title: Kurrama River\\
Snippet: The Kurrama River, or Kurram River, originates from the watershed of Spin Ghar region in the Paktia province of Afghanistan and the Kurram District of Pakistan. ...\\
\#4 Answerbox Title: Kurram Garhi\\
Snippet: Kurram Garhi is a small village located near the city of Bannu, which is the part of Khyber Pakhtunkhwa province of Pakistan.\\
Question: When did the director of film Hypocrite (Film) die?\\
Answer: The film Hypocrite was directed by Miguel Morayta. Miguel Morayta died on 19 June 2013. So the answer is **19 June 2013**.\newline

......}
    \caption{Demonstrations of open-book reasoning (SERP) on 2WikiMQA dataset. (5 shot)}
    \label{demo:reason_serp}
\end{figure*}

\end{document}